\pdfoutput=1
\documentclass[11pt]{article}

\usepackage[]{acl}

\usepackage{times}
\usepackage{latexsym}

\usepackage[T1]{fontenc}
\usepackage[utf8]{inputenc}

\usepackage{microtype}

\usepackage{amsfonts}
\usepackage{multirow}
\usepackage{graphicx}

\graphicspath{{pictures/}}

\title{BOS at LSCDiscovery: Lexical Substitution for Interpretable Lexical Semantic Change Detection}

\author{Artem Kudisov$^{\nabla}$ \\
  \And Nikolay Arefyev$^{\diamondsuit,\nabla,\triangle}$ \\
  \AND
  \vspace{-0.8cm}\\
  $^{\nabla}$Lomonosov Moscow State University / Moscow, Russia \\
  $^{\triangle}$National Research University Higher School of Economics / Moscow, Russia \\
  $^\diamondsuit$Samsung Research Center Russia / Moscow, Russia \\
  \texttt{dark.artbeam@gmail.com,} \texttt{nick.arefyev@gmail.com} \\
}

\begin{document}
\maketitle
\begin{abstract}
We propose a solution for the LSCDiscovery shared task on Lexical Semantic Change Detection in Spanish. Our approach is based on generating lexical substitutes that describe old and new senses of a given word. This approach achieves the second best result in sense loss and sense gain detection subtasks. By observing those substitutes that are specific for only one time period, one can understand which senses were obtained or lost. This allows providing more detailed information about semantic change to the user and makes our method interpretable. 
\end{abstract}

\section{Introduction}

LSCDiscovery is a shared task on Lexical Semantic Change Detection (LSCD) in Spanish~\cite{lscd2022}. The participants were provided with two corpora in Spanish, corresponding to 1810-1906 and 1994-2020 respectively, and were asked to solve two subtasks. In the first subtask the participants were asked to rank the given list of about 4K words according to the degree of their semantic change. The second subtask required to determine for each given word if its senses occurring in two corpora are different (and optionally, if it has acquired some new senses, and if it has lost any old ones). 

\section{Background}

Our approach is based on the bag-of-substitutes (BOS) representation of word meaning in context~\cite{baskaya-etal-2013-ai,arefyev-zhikov-2020-bos}. Lexical substitutes are those words that can replace a given target word in a given text fragment without making this fragment ungrammatical or substantially changing the meaning of the target word. For ambiguous words, lexical substitutes depend on their meaning expressed in a particular context. For instance, some reasonable substitutes for the word \textit{fly} in the sentence \textit{A noisy fly sat on my shoulder} are \textit{bug}, \textit{beetle}, \textit{butterfly}, \textit{firefly}, \textit{insect}, etc. But in the sentence \textit{We will fly to London} they are different: \textit{walk}, \textit{run}, \textit{bike}, etc. 

In order to generate lexical substitutes, we employ the XLM-R\footnote{The pre-trained xlmr.large from \href{https://github.com/pytorch/fairseq/blob/main/examples/xlmr/README.md}{fairseq} library is used without any fine-tuning.} masked language model~\cite{xlmr}. This model was pre-trained on 2.5T of data in 100 languages as a masked language model, i.e. it received text fragments with some tokens hidden (replaced with the special \texttt{<mask>} token) and was trained to guess those hidden tokens by their context. This kind of pre-training is partially aligned with the lexical substitution task because the model can predict words compatible with the given context. However, there are no guarantees that these words are similar or related by meaning to the target word.
Suitable types of lexical substitutes (e.g., synonyms, hypernyms, co-hyponyms) and suitable degree of their similarity to the target word depend on the target task and can be controlled with various techniques explored in~\cite{keeptarget2020}. In our solution, we employ the dynamic patterns proposed by~\citet{amrami2018} and explained in~\ref{sec:subst_gen}.

\begin{table*}[t]
    \centering
    \small
    \begin{tabular}{l|c|l|c|l|c}
    \hline
        \multicolumn{2}{c|}{\texttt{<mask>}-(y-\texttt{T})} & \multicolumn{2}{c|}{\texttt{<mask><mask>}-(y-\texttt{T})} &
        \multicolumn{2}{c}{\texttt{<mask>}-(incluso-\texttt{T})}\\
    \hline
        Substitute & Prob. & Substitute & Prob. & Substitute & Prob.\\
    \hline
        documentos (documents) & 0.367 & archivos (records) & 0.016 & documentos (documents) & 0.391\\
        libros (books) & 0.160 & escritos (letters) & 0.012 & libros (books) & 0.082\\
        datos (data) & 0.052  & informes (reports) & 0.010 & datos (data) & 0.039\\
        actos (acts) & 0.036 & \parbox[c]{3cm}{dos documentos (two documents)} & 0.010 & textos (texts) & 0.037 \\
        textos (texts) & 0.032 & expedientes (records) & 0.008 & contratos (contracts) & 0.014\\
    \hline
    \end{tabular}
    \caption{For the word \textit{actas} (\textit{reports}) in \textit{ayer recibimos dos actas literales} (\textit{yesterday we received two verbatim reports}), 5 most probable substitutes with 1 or 2 subwords are shown. The patterns with \textit{y} (\textit{and}) and \textit{incluso} (\textit{including}).}
    \label{tab:substs1}
\end{table*}

Unlike the traditional bag-of-words representation, which contains those words that occur in a text fragment, the BOS representation is built from lexical substitutes. Thus, it better represents the meaning of some specific target word in a given text fragment rather than the whole fragment in general. Clustering of the BOS vectors is a successful approach to solve the Word Sense Induction (WSI) task, i.e. to discover senses of ambiguous words. This approach was explored in many papers, including~\cite{baskaya-etal-2013-ai,amrami2018,amrami2019,Arefyev_2019,keeptarget2020} among others. Also, a substitution-based WSI model was employed to solve the LSCD task in~\cite{arefyev-zhikov-2020-bos,dialog2021-bykov}. However, in our solution we avoid solving the more general and probably more difficult WSI task that requires clustering. Instead, we propose methods to directly obtain LSCD predictions from the BOS vectors. 

\section{Model description}

For each target word we sample some examples of its usage from both corpora and generate lexical substitutes for them. Then we build two sets of BOS vectors for old and new examples, describing old and new senses of the word respectively. Finally, the distances from old to new examples are calculated, and their average is returned as the predicted score of graded change. Following previous works on LSCD~\cite{giulianelli-etal-2020-analysing,laicher-etal-2021-explaining}, we will denote this average as the Average Pairwise Distance (APD). Notice that our vector representation is very different from those works.

For the second subtask, if APD is greater than a certain threshold, we predict that this word has changed its meaning. To determine whether it has acquired new senses and whether any old senses were lost, we propose three different methods based on pairwise distances.

\subsection{Collected data}

For each target word $w_i$, we lemmatize\footnote{We used the Spanish lemmatizer from Spacy proposed by the organizers.} both corpora and retrieve all examples with $w_i$ in different grammatical forms. Then we take the same number $N_i$ of examples from the old and the modern set of examples.\footnote{If possible, $N_i=100$ examples are sampled without replacement from each set. Otherwise, we take all $N_i<100$ examples from the smaller set and sample the same number of examples from another set.}

\subsection{Substitute generation}
\label{sec:subst_gen}
For each example we generate several types of substitutes with different dynamic patterns, post-process them and combine together to get a single vector representation. Dynamic patterns are similar to the Hearst patterns by nature~\cite{hearst}. They were proposed in~\cite{amrami2018} to obtain from masked language models those substitutes that do not only fit the given context, but also are similar or related to the target word by meaning. For instance, using patterns with the Spanish conjunction \textit{y} (English: \textit{and}) we hope to obtain mostly co-hyponyms of the target word, while patterns with the adverb \textit{incluso} (English: \textit{including}) shall bias the model towards generating hypernyms or hyponyms, depending on the position of the target word. Table~\ref{tab:substs1} shows some examples.

\begin{table}[t]
    \centering
    \small
    \begin{tabular}{l|c}
    \hline
         Pattern &  weight \\
    \hline
        \textit{\texttt{<mask>}} & 0.25 \\
        \textit{\texttt{<mask>}-(y-\texttt{T})} & 0.25 \\
        \textit{\texttt{T}-(y-\texttt{<mask>})} & 0.25 \\
        \textit{\texttt{<mask>}-(incluso-\texttt{T})} & 0.0625 \\
        \textit{\texttt{T}-(incluso-\texttt{<mask>})} & 0.0625 \\
        \textit{\texttt{<mask>}-(por-ejemplo-\texttt{T})} & 0.0625 \\
        \textit{\texttt{T}-(por-ejemplo-\texttt{<mask>})} & 0.0625 \\
    \hline
    \end{tabular}
    \caption{In LS\_m1\_7, we employ 7 single-subword patterns with \textit{y} (\textit{and}), \textit{incluso} (\textit{including}) and \textit{por ejemplo} (\textit{for example}) with the specified weights.}
    \label{tab:patterns1}
\end{table}

\begin{table}[t]
    \centering
    \small
    \begin{tabular}{|l|l|c|}
    \hline
         LS\_m1\_2 patterns & LS\_m2\_2 patterns &  weight \\
    \hline
        \textit{\texttt{<mask>}-(y-\texttt{T})}  & \textit{\texttt{<mask><mask>}-(y-\texttt{T})} & 0.5\\
        \textit{\texttt{T}-(y-\texttt{<mask>})} & \textit{\texttt{<mask><mask>}-(y-\texttt{T})} & 0.5\\
    \hline
    \end{tabular}
    \caption{In LS\_m1\_2 and LS\_m2\_2 we employ 2 single-subword and 2 two-subword patterns respectively.}
    \label{tab:patterns2}
\end{table}

Table~\ref{tab:patterns1} lists all dynamic patterns we use. All patterns contain the special token \texttt{<mask>} that XLM-R is asked to recover, and some of them contain the variable \texttt{T} representing the target word. Given a pattern and an example for some target word, first we replace the target word with this pattern, and then replace the variable \texttt{T} (if any) back with the target word. For simplicity, let us consider an example in English. Given the sentence \textit{We can fly to London} and using the pattern \textit{\texttt{<mask>} (and \texttt{T})}, we first obtain \textit{We can \texttt{<mask>} (and \texttt{T}) to London}, and finally have \textit{We can \texttt{<mask>} (and fly) to London}. 

The vocabulary of XLM-R consists of 250K subwords in 100 different languages, which are sometimes whole frequent words, but most often pieces of words. To better describe word meaning, we generate substitutes consisting of different number of subwords. To achieve this, we apply patterns with several \texttt{<mask>} tokens, for instance,\textit{\texttt{<mask><mask>} (y \texttt{T})}. 

To find probable sequences of subwords that could fill the \texttt{<mask>} tokens, we apply a slightly modified greedy decoding strategy. For the leftmost \texttt{<mask>} token, $topK=150$ most probable subwords are predicted first. Then for each of those subwords we  generate one continuation using greedy decoding. Below we will say that a substitute is not generated for a particular pattern in a particular example if it was not among $topK$ substitutes generated this way. For computational reasons, we generated only substitutes with one or two subwords and did not apply beam search for decoding. Examples of two-subword substitutes are in table~\ref{tab:substs1}.

\subsection{Substitute post-processing and combination}
\label{sec:subst_postprocess}

Next, we post-process all substitutes for each example: convert them to lower case, remove all words except for the last one from multi-word substitutes, apply stemming.\footnote{The Spanish stemming from \href{https://www.nltk.org/api/nltk.stem.snowball.html}{nltk.stem.snowball}  was used.} After post-processing, we sum the probabilities of duplicated substitutes.

For each example, we combine substitutes generated for different patterns by calculating the weighted average of the corresponding probability distributions. In \textbf{LS\_m1} and \textbf{LS\_m2} (Lexical Substitution with one-subword substitutes and two-subword substitutes respectively), for combination we use patterns and weights presented in Tables~\ref{tab:patterns1} and~\ref{tab:patterns2}. The weights were selected based on a few experiments on the development set consisting of 20 words, so these weights are likely suboptimal.
% After we get rid of duplicated substitutes by summing their probabilities again. 
It is possible that one of the substitutes is not generated by XLM-R for a certain pattern. In this case, during combination we assume that the corresponding probability is equal to the minimal probability among all substitutes generated for this pattern. 

\subsection{BOS vectors}

For each target word $w_i$ we build $2N_i$ BOS vectors for old and new examples. These vectors are basically bag-of-word vectors built for $topK$ most probable substitutes for each example. Only substitutes that were generated for more than 3\% and less than 90\% of examples of the target word are taken into account\footnote{We used \href{https://scikit-learn.org/stable/modules/generated/sklearn.feature_extraction.text.CountVectorizer.html}{CountVectorizer} from scikit-learn, where $min\_df=0.03$ was selected in range from 0 to 0.05 with 0.01 step and $max\_df=0.9$ was selected in range from 0.85 to 1 with 0.01 step.}.

\subsection{Graded Change Discovery}

\textbf{APD (Average Pairwise Distance)}. After building the BOS vectors, we calculate the cosine distance from each old to each new example, resulting in a matrix of size ${N_i\times N_i}$. The APD is calculated by averaging all cells in this matrix. Finally, we sort test words according to their APDs and submit their ranks as the predicted change scores.\footnote{There was a mistake in the original implementation of the ranking procedure. After the competition we fixed it, which significantly improved the results of this method (see table~\ref{tab:test_phase1} for comparison).}

\begin{table}[t]
    \centering
    \small
    \begin{tabular}{|l|c|c|}
    \hline
         \textbf{Model/Team} & \textbf{JSD,SPR} & \textbf{COMPARE,SPR}\\
    \hline
        \multicolumn{3}{|c|}{\textbf{baselines}} \\
    \hline
        baseline1 & \textbf{0.543} (4) & \textbf{0.561} (3)\\
        baseline2 & 0.092 (8) & 0.088 (6)\\ [0.5ex]
    \hline
    \hline
        \multicolumn{3}{|c|}{\textbf{best results of other teams}} \\
    \hline
        myrachins & \textbf{0.735} (1) & \textbf{0.842} (1)\\
        UsrD7 & 0.702 (2) & 0.829 (2) \\ 
        aishein & 0.553 (3) & 0.558 (4)\\ [0.5ex]
    \hline
    \hline
        \multicolumn{3}{|c|}{\textbf{our results}} \\
    \hline
        $^\#$LS\_m1\_7+APD & -0.125 (9) & -0.129 (8)\\ [0.5ex]
    \hline
    \hline
        \multicolumn{3}{|c|}{\textbf{our post-evaluation results}} \\
    \hline
        LS\_m1\_7+APD & \textbf{0.584} (3*) & 0.598 (3*)\\ 
        LS\_m1\_2+APD & 0.562 (3*) & 0.562 (3*)\\ 
        LS\_m2\_2+APD & 0.576 (3*) & \textbf{0.637} (3*)\\
    \hline
    \end{tabular}
    
    \caption{Graded Change Discovery results. $\#$ denotes the buggy implementation. * denotes possible ranks of the corresponding results in the leaderboard.} 
    \label{tab:test_phase1}
\end{table}

\begin{table}[t]
    \centering
    \footnotesize
    \begin{tabular}{|l|ccc|}
    \hline
         \textbf{Model/Team} & \textbf{CH, F1} & \textbf{GAIN, F1} & \textbf{LOSS, F1} \\
    \hline
        \multicolumn{4}{|c|}{\textbf{baselines}}\\
    \hline
        baseline1 & 0.537 (9) & NaN (8) & NaN (6) \\
        baseline2 & 0.222 (10) & 0.211 (7) & 0.000 (6) \\ [0.5ex]
    \hline
    \hline
        \multicolumn{4}{|c|}{\textbf{best results of other teams}}\\
    \hline
        myrachins & \textbf{0.716} (1) & \textbf{0.491} (3) & \textbf{0.688} (1) \\
        dteodore & 0.709 (2) & 0.000 (8) & 0.000 (6)\\ 
        rombek & 0.687 (3) & 0.490 (4) & 0.593 (3) \\ [0.5ex]
    \hline
    \hline
        \multicolumn{4}{|c|}{\textbf{our results}}\\
    \hline
         LS\_m1\_7+AID & \textbf{0.658} (4*) & 0.393 (6*) & 0.137 (6*) \\ 
         LS\_m2\_2+min & 0.636 (6*) & 0.418 (6*) & \textbf{0.610} (2*) \\ 
         LS\_m1\_7+perc. & \textbf{0.658} (4) & \textbf{0.520} (2) & 0.600 (2) \\ [0.5ex]
    \hline
    \hline
        \multicolumn{4}{|c|}{\textbf{our post-evaluation results}}\\
    \hline
         LS\_m1\_2+AID & 0.628 (7*) & 0.4 (6*) & 0.076 (6*) \\ 
         LS\_m1\_2+min & 0.628 (7*) & \textbf{0.583} (2*) & 0.387 (5*) \\ 
         LS\_m1\_2+perc.& 0.628 (7*) & 0.486 (5*) & \textbf{0.608} (2*) \\
     \hline
         LS\_m2\_2+AID & 0.636 (6*) & 0.382 (6*) & 0.193 (6*) \\ 
         LS\_m2\_2+perc. & 0.636 (6*)& 0.376 (6*) & 0.600 (2*) \\ 
        \hline
         LS\_m1\_7+min & \textbf{0.658} (4*) &  0.533 (2*) &  0.564 (5*) \\ 
     \hline
    \end{tabular}
    
    \caption{Binary Change Detection results. * denotes possible ranks of the corresponding results in the leaderboard.}
    \label{tab:test_phase2}
\end{table}

\subsection{Binary Change Detection}

For the main Binary Change Detection subtask, if the calculated APD is greater than the certain $threshold$\footnote{$threshold=0.8$ was selected on the development set in the range from 0.7 to 0.9 with 0.05 step.}, then we predict that this word has changed its meaning. In this case we also try to determine if it has acquired new senses and if it has lost some old ones (sense loss and sense gain detection subtasks). We try three methods to determine that. 

\textbf{AID (Average Inner Distance)}. We calculate APDs between only new examples $AID_1$ and between only old examples $AID_2$. If $AID_1 > (AID_2 - b_1)$, we predict that a new sense appeared. If $AID_2 > (AID_1 - b_2)$, we predict that an old sense is lost.\footnote{$b_1 = 0.03$, $b_2=-0.03$. These values were selected on the development set in the range from -0.1 to 0.1 with 0.01 step.} Thus, we assume that a difference in average inner distances for two sets of examples indicates that there is a difference in underlying sets of senses.

\textbf{min}. We calculate an ${N_i\times N_i}$ matrix of pairwise distances from old to new examples and assume that if some new sense appeared, then a new example exists that is far from all old examples. Thus, if there is at least one new example whose minimal distance to the old examples is greater than some $threshold$ \footnote{$threshold=0.8$ was selected on the development set in the range from 0.7 to 0.9 with 0.05 step.}, we predict that a new sense appeared. Sense loss is determined symmetrically.

\textbf{perc. (percentile)}. This is similar to the previous method, but we calculate the 5th percentile instead of the minimum, i.e. we allow at most 5\% of examples from the old corpus to be closer to an example of the new sense from the new corpus than the specified threshold. We assume that this should make the model less sensitive to noisy examples and more stable.

\begin{table}[t]
    \centering
    \footnotesize
    \begin{tabular}{|l|cc|}
    \hline
        Model & GAIN,F1 & LOSS,F1 \\
    \hline
        LS\_m1\_2+AID &  0.4 (6*) & 0.076 (6*) \\ 
        LS\_m1\_2+min &  \textbf{0.583} (2*) & 0.387 (5*) \\ 
        LS\_m1\_2+perc.&  0.486 (5*) & 0.608 (2*) \\
    \hline
        LS\_m2\_2+AID & 0.382 (6*) & 0.193 (6*) \\ 
        LS\_m2\_2+min &  0.418 (6*) & \textbf{0.610} (2*)\\
        LS\_m2\_2+perc. &  0.376 (6*) & 0.600 (2*) \\ 
        \hline
        LS\_m1\_7+AID &  0.393 (6*) & 0.137 (6*)\\
        LS\_m1\_7+min & 0.533 (2*) &  0.564 (5*) \\ 
        LS\_m1\_7+perc. &  0.520 (2) & 0.600 (2)\\
    \hline
    \end{tabular}
    
    \caption{Comparison of aggregation methods. * denotes possible ranks of the corresponding results in the leaderboard.}
    \label{tab:aggregation_comparison}
\end{table}

\section{Experiments and Results}

\subsection{Phase 1: Graded Change Discovery}

In this subtask, it was required to rank about 4K target words according to their degree of semantic change (the higher rank, the stronger change). The final quality of ranking was evaluated for 60 hidden words only by the Spearman’s correlation with the gold ranks~\cite{bolboaca2006pearson}.

Table~\ref{tab:test_phase1} provides the results for the first phase. Our original implementation of the ranking procedure had mistakes in the ranking procedure, so the results are poor. After the competition, we fixed the mistake and obtained the correct results, which are comparable to the 3rd best participant in the leaderboard. 

\textbf{LS\_m1\_2} and \textbf{LS\_m2\_2} differ only in the number of masks in the used patterns. So comparing their scores, we can say that using two-subword substitutes is more preferable than one-subword substitutes. In \textbf{LS\_m2\_7} seven patterns are combined compared to two patters in \textbf{LS\_m1\_2}, this gives a significant improvement despite somewhat arbitrarily selected weights. Developing some principled ways of finding promising dynamic patterns and weights for their combination is a reasonable direction for future work. \textbf{LS\_m1\_7} has a slightly higher JSD,SPR score, but its COMPARE,SPR score is lower and it uses a more complex pattern combination than \textbf{LS\_m2\_2}. A more detailed investigation is presented in Appendix~\ref{appendix:substs_analysis}.

\subsection{Phase 2: Binary Change Detection}

In this subtask the participants were asked to determine if target words have changed their meanings. And if so, how exactly (have acquired and/or have lost senses). Three F1-scores are calculated: Binary Change Detection (CH), Sense Gain Detection (GAIN), Sense Loss Detection (LOSS). Results are presented in Table~\ref{tab:test_phase2} where we have the 2nd best submission for GAIN and LOSS optional subtasks.

\textbf{LS\_m1\_2 + APD} and \textbf{LS\_m2\_m2 + APD} have $0.628$ and $0.636$ CH,F1 scores respectively, which means that using two-subword substitutes is slightly better than one-subword. But in the case of \textbf{LS\_m1\_7 + APD} we already get $0.658$ CH,F1 resulting in the 4th rank.

Using \textbf{AID} method does result in good GAIN,F1 and LOSS,F1 scores (Table~\ref{tab:aggregation_comparison}). At the same time \textbf{min} and \textbf{percentile} show a better results but they highly depend on used LS patterns, i.e., in the some cases these methods improves only GAIN,F1 or LOSS,F1 scores, but not both of them.

%But \textbf{LS\_m1\_7 + APD + percentile} has much higher LOSS,F1 and GAIN,F1 resulting in the 2nd best submission for these optional subtasks.

\section{Discriminative substitutes}

The main advantage of LS-based models is their interpretability. We can roughly understand word meanings looking at the discriminative substitutes, i.e. the substitutes specific for a particular subset of examples.

Table~\ref{tab:discriminative} provides some examples for \textit{disco} (\textit{disc}) and \textit{satélite} (\textit{satellite}). We take old examples $O$ and those new examples $M$, that were determined by LS\_m1+percentile model as being far away from $O$. Then we find substitutes with the largest ratio $\frac{P(w|M)}{P(w|O)}$\footnote{If the word denominator is 0, we demand $P(w|M)$ to be greater 0.2, otherwise we don't consider such word.}, i.e. those substitutes that are rarely generated for old examples but frequently generated for new examples that are not similar to any old examples.

\begin{table}[h]
    \centering
    \small
    \begin{tabular}{cl|cl}
    \hline
         \multicolumn{2}{c|}{Disco (disc)} & \multicolumn{2}{c}{Satélite (satellite)}\\
    \hline
        LP & 0.72/0.00 & CD & 1.00/0.00\\
        EP & 0.55/0.00 &  video & 1.00/0.00\\
        documentos & 0.55/0.00 &  internet & 1.00/0.00\\
        videos & 0.50/0.00 &  televisión & 0.88/0.00\\
        mp & 0.50/0.00 &  FM & 0.88/0.00\\
        anime & 0.44/0.00 & señal & 0.88/0.00\\
        memoria & 0.44/0.00 & Internet & 0.88/0.00\\
        PC & 0.44/0.00 & canal & 0.88/0.00 \\
        USB & 0.44/0.00 & TV & 0.88/0.00\\
        b & 0.44/0.00 & web & 0.88/0.00\\
        MP & 0.44/0.00 & vídeo & 0.77/0.00\\
    \hline
    \end{tabular}
    \caption{Discriminative substitutes generated for the \textit{\texttt{<mask>} (y \texttt{T})} pattern. The probabilities $P(w|M)$ and $P(w|O)$ are shown for each substitute.  Documentos is 'documents', señal is 'signal', memoria is 'memory' and canal is 'channel'.} 
    \label{tab:discriminative}
\end{table}

From the Table~\ref{tab:discriminative} we can see that \textit{disco} (\textit{disc}) and \textit{satélite} (\textit{satellite}) have acquired new senses as \textit{a data storage device} and \textit{satellite television} respectively.

\section{Efficiency}

The set of the target words proposed in Phase 1 was supposed to be a challenge for participants due to its size. For 4385 words given we have collected about 777K examples. Generation of substitutes for all examples took 13 GPU-hours and 310 GPU-hours for each one-mask and two-mask pattern respectively on V100 GPUs. All other steps took incomparably less time.

\section{Conclusion}

We have proposed an interpretable approach to lexical semantic change detection. This approach shows the 2nd best result for sense loss and sense gain detection subtasks. It provides techniques to understand which senses were obtained or lost by a word.

\section*{Acknowledgements}
We are grateful to our anonymous reviewers. This research was partially supported by the Basic Research Program at the HSE University and through computational resources of HPC facilities at HSE University~\cite{Kostenetskiy_2021}.

\bibliography{anthology,custom}
\bibliographystyle{acl_natbib}

\newpage

\appendix

\section{Substitute analysis}
\label{appendix:substs_analysis}

Our models mostly depend on the used LS patterns and ways of their combination. So it is important to make some investigations about them. In this section we study the following questions. 
\begin{itemize}
    \item Which single-subword pattern gives the best results and how these results depend on the number of substitutes generate (topk)?
    \item Is it better to use single-subword or multi-subword substitutes?
    \item Do brackets and dashes affect the results?
\end{itemize}

\begin{figure}[h]
    \includegraphics[width=\columnwidth]{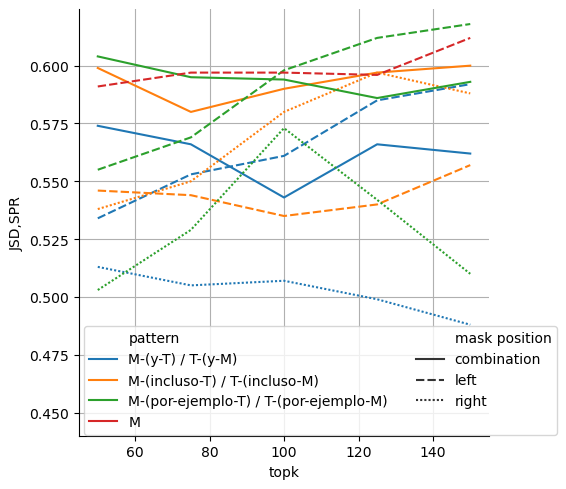}
    \caption[width=\columnwidth]{Dependence of the JSD,SPR score on the pattern and topk.}
    \label{fig:compare_1}
\end{figure}

\begin{figure}[h]
    \includegraphics[width=\columnwidth]{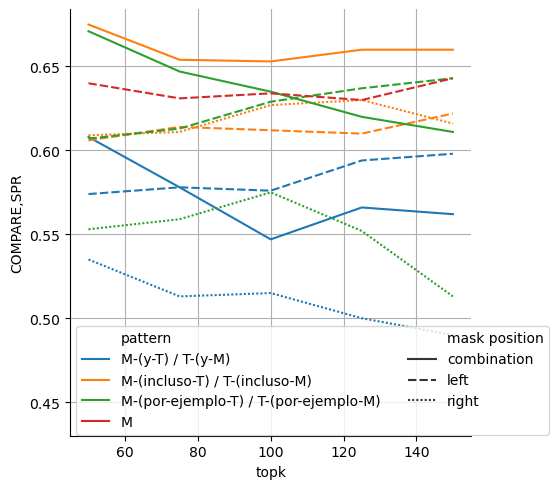}
    \caption[width=\columnwidth]{Dependence of the COMPARE,SPR score on the pattern and topk.}
    \label{fig:compare_2}
\end{figure}

For brevity, we will use \texttt{M} instead of \texttt{<mask>} in the pattern descriptions. In the following figures \texttt{mask position} describes the position of the \texttt{<mask>} token. For example, if the pattern is \textit{\texttt{M} (y \texttt{T}) / \texttt{T} (y \texttt{M})}, \texttt{mask position}=\texttt{left} refers to the pattern \textit{\texttt{M} (y \texttt{T})}, and \texttt{mask position}=\texttt{right} refers to \textit{\texttt{T} (y \texttt{M})}. Finally, \texttt{mask position}=\texttt{combination} denotes the combination of these patterns with equal weights.

%For convenience, in the all following figures we use special notations to describe used patterns. Thus, \texttt{conj} describes which Spanish conjunction was used in the pattern and \texttt{type} describes the certain type of the pattern. For example, \texttt{conj}=\texttt{incluso} with \texttt{type}=\texttt{right} denotes \textit{\texttt{T} (incluso \texttt{<mask>}}) and \texttt{conj}=\texttt{y} with \texttt{type}=\texttt{combination} denotes the combination of \textit{\texttt{T} (y \texttt{<mask>}}) and \textit{\texttt{<mask>} (y \texttt{T}}) with equal weights. \textit{\texttt{<mask>}} has empty \texttt{conj} and \texttt{type}=\texttt{symmetric}.

\subsection{Single-subword patterns}

In \textbf{LS\_m1\_7} we use 7 patterns with different weights, which were selected after only a few experiments on the development set. In this section we study how the results depend on the patterns and try to find simpler and more intuitive ways of the substitute combination. Figures~\ref{fig:compare_1} and~\ref{fig:compare_2} show JSD,SPR and COMPARE,SPR for different patterns.

It is interesting that in all cases the \texttt{left} patterns give better results than the \texttt{right} ones, except for the incluso-based patterns. Also in all cases the combination averages the results of both patterns, again except for the combination of incluso-based patterns which on the contrary improves the results. 

\begin{figure}[h]
    \includegraphics[width=\columnwidth]{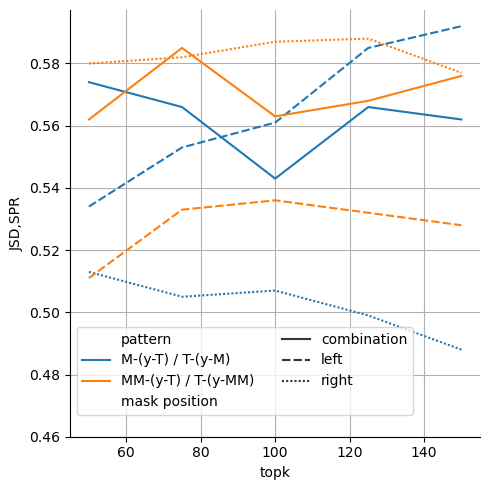}
    \caption{Comparison of one-subword and two-subword substitutes.}
    \label{fig:1vs2_1}
\end{figure}

\subsection{One-subword substitutes vs. two-subword substitutes}

We assume that using more masks should improves results because this allows to generate more diverse substitutes. Figure~\ref{fig:1vs2_1} provides comparison of patterns with different number of masks. As we suspect, using \textit{\texttt{T} (y \texttt{MM})} pattern gives a much better results than \textit{\texttt{T} (y \texttt{M})}. However combination of two-mask patterns results in just slightly higher score and one-mask pattern \textit{\texttt{M} (y \texttt{T})} even outperforms \textit{\texttt{MM} (y \texttt{T})}.

\begin{figure}[h]
    \includegraphics[width=\columnwidth]{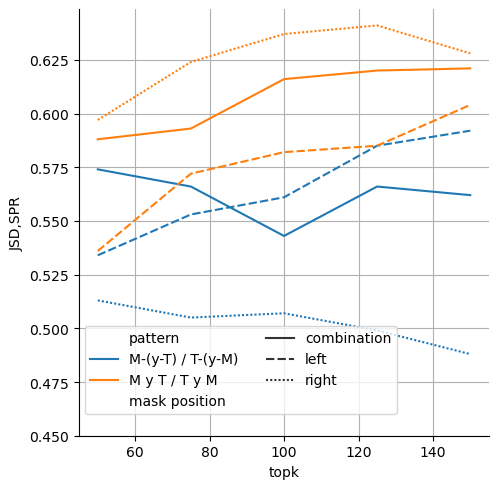}
    \caption[width=\columnwidth]{Comparison patterns with and without brackets.}
    \label{fig:brackets_1}
\end{figure}

 \begin{figure}[h]
    \includegraphics[width=\columnwidth]{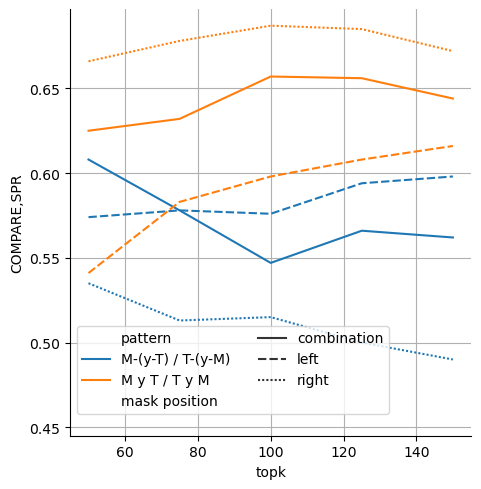}
    \caption[width=\columnwidth]{Comparison patterns with and without brackets.}
    \label{fig:brackets_2}
\end{figure}

\subsection{Patterns without brackets and dashes}

In the patterns discussed above we have extra dashes which were added by mistake and potentially could affect the results, so firstly we remove them from patterns. Also we have assumption that using brackets is not common thing in Spanish so such patterns could spoil generated substitutes and final results. To prove it we decide to compare \texttt{y}-based patterns with and without brackets and dashes. 

In the Figures~\ref{fig:brackets_1} and~\ref{fig:brackets_2} we can see that in all cases refusal to use brackets and dashes improves our results quite well, especially the \texttt{right} pattern get around 0.1 growth in JSD,SPR and COMPARE,SPR scores.

\end{document}